\documentclass[10pt,twocolumn,letterpaper]{article}

\usepackage{multirow}
\usepackage{cvpr}
\usepackage{times}
\usepackage{epsfig}
\usepackage{graphicx}
\usepackage{amsmath}
\usepackage{amssymb}
\usepackage[export]{adjustbox}
\usepackage{siunitx}
\usepackage[numbers,sort]{natbib} 

\usepackage[pagebackref=true,breaklinks=true,letterpaper=true,colorlinks,bookmarks=false]{hyperref}

\cvprfinalcopy 

\def\cvprPaperID{2790} 

\ifcvprfinal\fi
\begin{document}

\title{Quo Vadis, Action Recognition? A New Model and the Kinetics Dataset}

\author{Jo{\~a}o~Carreira$^\dagger$ \\
{\tt\small joaoluis@google.com}
\and
Andrew~Zisserman$^{\dagger,*}$ \\
{\tt\small zisserman@google.com}
\and
$^\dagger$DeepMind
\quad\quad\quad
$^*$Department of Engineering Science, University of Oxford\\
}

\maketitle

\begin{abstract}

The paucity of videos in current action classification datasets
(UCF-101 and HMDB-51) has made it difficult to identify good video
architectures, as most methods obtain similar performance on existing
small-scale benchmarks. This paper re-evaluates state-of-the-art
architectures in light of the new Kinetics Human Action Video 
dataset. Kinetics has two orders of magnitude
more data, with 400 human action classes and over 400 clips per class,
and is collected from
realistic, challenging YouTube videos. We provide an analysis on how
current architectures fare on the task of action classification on
this dataset and how much performance improves on the smaller 
benchmark datasets
after pre-training on Kinetics.

We also introduce a new Two-Stream Inflated 3D ConvNet (I3D) that is
based on 2D ConvNet \textit{inflation}: filters and pooling kernels of
very deep image classification ConvNets are expanded into 3D, making
it possible to learn seamless spatio-temporal feature extractors from
video while leveraging successful ImageNet architecture designs and
even their parameters. We show that, after pre-training on Kinetics,
I3D models considerably improve upon the state-of-the-art in action
classification, reaching 80.9\% on HMDB-51 and 98.0\% on UCF-101.
\end{abstract}

\section{Introduction}

\begin{figure}
  \centering
    \includegraphics[width=0.45\textwidth]{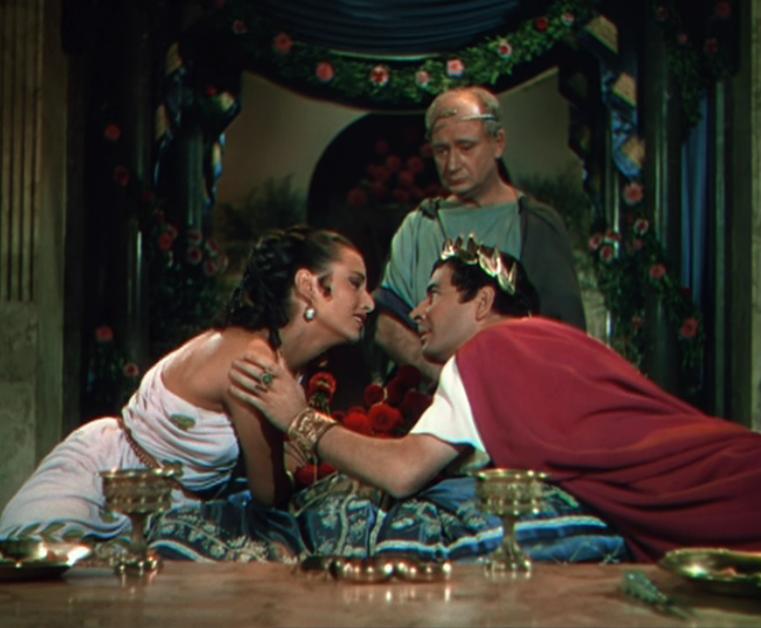}
    \caption{{\bf A still from `Quo Vadis' (1951).} Where is this going? Are these actors about to kiss each other, or have they just done so? More importantly, where is action recognition going? Actions can be ambiguous in individual frames, but the limitations of existing action recognition datasets has meant that the best-performing video architectures do not depart significantly from single-image analysis, where they rely on powerful image classifiers trained on ImageNet. In this paper we demonstrate that video models are best pre-trained on videos and report significant improvements by using spatio-temporal classifiers pre-trained on Kinetics, a freshly collected, large, challenging human action video dataset.}
    \label{fig:teaser}
\end{figure}

One of the unexpected benefits of the ImageNet challenge has been the
discovery that deep architectures trained on the 1000 images of 1000
categories, can be used for other tasks and in other domains. One of
the early examples of this was using the fc7 features from a network
trained on ImageNet for the PASCAL VOC classification and detection
challenge~\cite{girshick2014rich,Oquab2014LearningAT}. Furthermore, improvements in the
deep architecture, changing from AlexNet to VGG-16, immediately fed
through to commensurate improvements in the PASCAL VOC
performance~\cite{ren2015faster}. Since then, there have been numerous
examples of ImageNet trained architectures warm starting or sufficing
entirely for other tasks, e.g.\ segmentation, depth prediction, pose
estimation, action classification.

In the video domain, it is an open question whether training an action
classification network on a sufficiently large dataset, will give a
similar boost in performance when applied to a different temporal task
or dataset.
The challenges of
building video datasets has meant that most popular benchmarks for
action recognition are small, having on the order of 10k videos.

In this paper we aim to provide an answer to this question using
the new Kinetics Human Action Video
Dataset~\cite{kinetics_arxiv}, which is two orders of magnitude
larger than previous datasets, HMDB-51~\cite{Kuehne11} and
UCF-101~\cite{soomro2012ucf101}. Kinetics has $400$ 
human action classes with more than $400$ examples for each class, each
from a unique YouTube video.
 
Our experimental strategy is to reimplement a number of representative
neural network architectures from the literature, and then analyze their transfer
behavior by first pre-training each one on Kinetics and then
fine-tuning each on HMDB-51 and UCF-101. The results suggest that there
is always a boost in performance by pre-training, but the extent of
the boost varies significantly with the type of architecture. Based on
these findings, we introduce a new model that has the capacity to take
advantage of pre-training on Kinetics, and can achieves a high
performance.  The model termed a ``Two-Stream Inflated 3D ConvNets''
(I3D), builds upon state-of-the-art image classification
architectures, but \textit{inflates} their filters and pooling kernels
(and optionally their parameters) into 3D, leading to very deep,
naturally spatio-temporal classifiers. An I3D model based on
Inception-v1~\cite{ioffe2015batch} obtains performance far exceeding
the state-of-the-art, after pre-training on Kinetics.

In our model comparisons, we did not consider more classic
approaches such as bag-of-visual-words
representations~\cite{wangICCV13,niebles2008unsupervised,laptev2008learning,fathi2008action}. 
However,
the Kinetics dataset is publicly available, so others can use it for
such comparisons.

The next section outlines the set of implemented action classification models.
Section~\ref{dataset} gives an overview of the Kinetics dataset.
Section~\ref{comp_arch} reports the performance of models on
previous benchmarks and on Kinetics, and section~\ref{comp_feat}
studies how well the features learned on Kinetics transfer to different
datasets. The paper concludes with a discussion of the results.

\section{Action Classification Architectures}
\label{sec:architectures}

While the development of image representation architectures has matured quickly in recent years, there is still no clear front running architecture for video. Some of the major differences in current video architectures are whether the convolutional and layers operators use 2D (image-based) or 3D (video-based) kernels; whether the input to the network is just an RGB video or it also includes pre-computed optical flow; and, in the case of 2D ConvNets, how information is propagated across frames, which can be done either using temporally-recurrent layers such as LSTMs, or feature aggregation over time.

In this paper we compare and study a subset of models that span most
of this space.  Among 2D ConvNet methods, we consider 
ConvNets
with LSTMs on top~\cite{yue2015beyond,donahue2015long}, and
two-stream
networks with two different types of stream
fusion~\cite{simonyan2014two,feichtenhofer2016convolutional}. We also
consider a 3D ConvNet~\cite{taylor2010convolutional,ji20133d}:
C3D \cite{tran2015learning}.

As the main technical contribution, we introduce Two-Stream Inflated
3D ConvNets (I3D). Due to the high-dimensionality of their
parameterization and the lack of labeled video data, previous 3D
ConvNets have been relatively shallow (up to 8 layers). Here we make
the observation that very deep image classification networks, 
such as Inception~\cite{ioffe2015batch}, 
VGG-16~\cite{simonyan2014very} and ResNet~\cite{he2016deep},
can be trivially inflated into spatio-temporal feature extractors,
and that their pre-trained weights provide a valuable
initialization. We also find that a two-stream configuration is still useful.

\begin{figure*}
  \centering
    \includegraphics[width=1.0\textwidth]{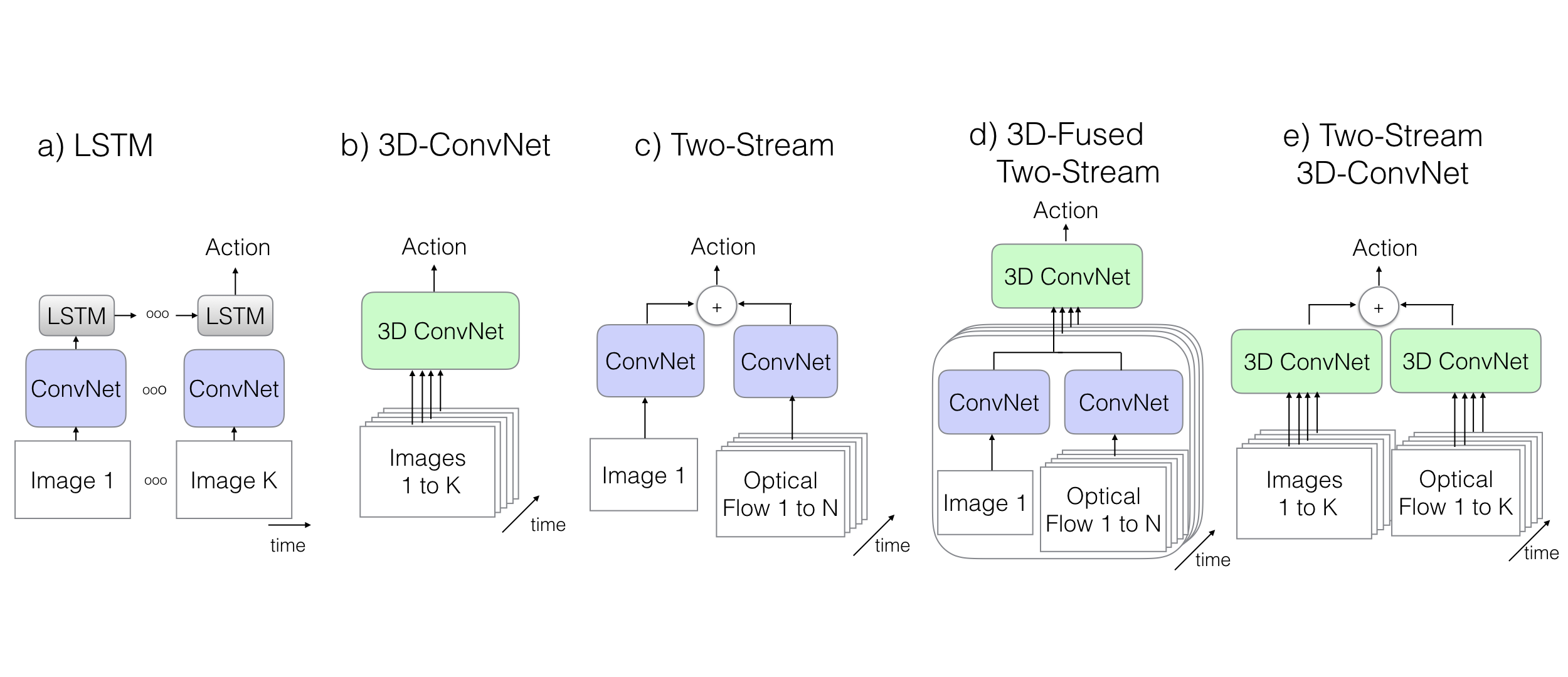}
    \caption{Video architectures considered in this paper. K stands for the total number of frames in a video, whereas N stands for a subset of neighboring frames of the video.}
    \label{fig:architectures}
\end{figure*}

A graphical overview of the five types of architectures we evaluate is shown in figure~\ref{fig:architectures} and the specification of their temporal interfaces is given in table \ref{tab:temporal_interfaces}.

Many of these models (all but C3D) have an Imagenet pre-trained model
as a subcomponent. Our experimental strategy assumes a common
ImageNet pre-trained image classification architecture as back bone,
and for this we chose Inception-v1 with batch
normalization~\cite{ioffe2015batch}, and morph it in different
ways. The expectation is that with this back bone in common, we will
be able to tease apart those changes that benefit action
classification the most.

\subsection{The Old I: ConvNet+LSTM}

The high performance of image classification networks makes it appealing to try to reuse them  with as minimal change as possible for video. This can be achieved by using them to extract features independently from each frame then pooling their predictions across the whole video \cite{karpathy2014large}. This is in the spirit of bag of words image modeling approaches \cite{wangICCV13,niebles2008unsupervised,laptev2008learning}; but while convenient in practice, it has the issue of entirely ignoring temporal structure (e.g.\ models can't potentially distinguish opening from closing a door). 

In theory, a more satisfying approach is to add a recurrent layer to the model \cite{yue2015beyond,donahue2015long}, such as an LSTM, which can encode state, and capture temporal ordering and long range dependencies. We position an LSTM layer with batch normalization (as proposed by Cooijmans {\it et al.}~\ \cite{cooijmans2016recurrent}) after the last average pooling layer of Inception-V1, with 512 hidden units. A fully connected layer is added on top for the classifier.

The model is trained using cross-entropy losses on the outputs at all time steps. During testing we consider only the output on the last frame. Input video frames are subsampled by keeping one out of every 5, from an original 25 frames-per-second stream. The full temporal footprint of all models is given in table~\ref{tab:temporal_interfaces}.

\subsection{The Old II: 3D ConvNets}

3D ConvNets seem like a natural approach to video modeling, and are
just like standard convolutional networks, but with spatio-temporal
filters. They have been explored several times, previously
\cite{taylor2010convolutional,ji20133d,tran2015learning, varol2017long}.  They have a
very important characteristic: they directly create hierarchical
representations of spatio-temporal data. One issue with these models
is that they have many more parameters than 2D ConvNets because of the
additional kernel dimension, and this makes them harder to train.
Also, they seem to preclude the benefits of ImageNet pre-training, and
consequently previous work has defined relatively shallow custom architectures and
trained them from scratch
\cite{taylor2010convolutional,ji20133d,tran2015learning,karpathy2014large}. Results
on benchmarks have shown promise but have not been competitive with
state-of-the-art, making this type of models a good candidate for
evaluation on our larger dataset.

For this paper we implemented a small variation of C3D
\cite{tran2015learning}, which has $8$ convolutional layers, $5$
pooling layers and $2$ fully connected layers at the top. The inputs
to the model are short $16$-frame clips with $112 \times 112$-pixel crops
as in the original implementation. 
Differently from~\cite{tran2015learning} we used batch
normalization after all convolutional and fully connected
layers. Another difference to the original model is in the first
pooling layer,   we use a temporal stride of $2$ instead of $1$,
which reduces the memory footprint and allows for bigger batches --
this was important for batch normalization (especially after the fully
connected layers, where there is no weight tying). Using this stride
we were able to train with 15 videos per batch per GPU using standard
K40 GPUs.

\subsection{The Old III: Two-Stream Networks}

LSTMs on features from the last layers of ConvNets can model high-level variation, but may not be able to capture fine low-level motion which is critical in many cases. It is also expensive to train as it requires unrolling the network through multiple frames for backpropagation-through-time. 

A different, very practical approach, introduced by Simonyan and Zisserman~\cite{simonyan2014two}, models short temporal snapshots of videos by averaging the predictions from a single RGB frame and a stack of $10$ externally computed optical flow frames, after passing them through two replicas of an ImageNet pre-trained ConvNet. The flow stream has an adapted input convolutional layer with twice as many input channels as flow frames (because flow has two channels, horizontal and vertical), and at test time multiple snapshots are sampled from the video and the action prediction is averaged. This was shown to get very high performance on existing benchmarks, while being very efficient to train and test.

A recent extension~\cite{feichtenhofer2016convolutional} fuses the
spatial and flow streams after the last network convolutional layer,
showing some improvement on HMDB while requiring less test time
augmentation (snapshot sampling). Our implementation follows this
paper approximately using Inception-V1. The inputs to the network are 5 consecutive RGB frames sampled 10 frames apart, as well as the
corresponding optical flow snippets.  The spatial and motion features
before the last average pooling layer of Inception-V1 ($5 \times 7 \times 7$ feature
grids, corresponding to time, x and y dimensions) are passed through a $3 \times 3 \times 3$ 3D convolutional layer with
512 output channels, followed by a $3 \times 3 \times 3$ 3D
max-pooling layer and through a final fully connected layer. The
weights of these new layers are initialized with Gaussian noise.

Both models, the original two-stream and the 3D fused version, are trained end-to-end (including the two-stream averaging process in the original model).

\subsection{The New: Two-Stream Inflated 3D ConvNets}

With this architecture, we show how 3D ConvNets can benefit from ImageNet
2D ConvNet designs and, optionally, from their learned parameters. We 
also adopt a two-stream configuration here -- it will be shown in
section~\ref{comp_arch}
that while 3D ConvNets can directly learn about temporal
patterns from an RGB stream, their performance can still be greatly
improved by including an optical-flow stream.

\vspace{3mm}
\noindent \textbf{Inflating 2D ConvNets into 3D.} 
A number of very successful image classification architectures have been 
developed over the years, in part through painstaking trial and
error. Instead of repeating the process for spatio-temporal models we
propose to simply convert successful image (2D) classification  models into 3D
ConvNets. This can be done by starting with a 2D architecture, and
\textit{inflating} all the filters and pooling kernels -- endowing
them with an additional temporal dimension. Filters are typically
square and we just make them cubic -- $N \times N$ filters become $N
\times N \times N$.

\vspace{3mm}
\noindent \textbf{Bootstrapping 3D filters from 2D Filters.} 
Besides the architecture, one may also want to bootstrap parameters
from the pre-trained ImageNet models. To do this, we observe that an
image can be converted into a (boring) video by copying it repeatedly
into a video sequence. The 3D models can then be implicitly pre-trained
on ImageNet, by satisfying what we call the boring-video fixed point:
the pooled activations on a boring video should be the same as on the
original single-image input. This can be achieved, thanks to
linearity, by repeating the weights of the 2D filters $N$ times
along the time dimension, and rescaling them by dividing by $N$. This
ensures that the convolutional filter response is the same. Since the
outputs of convolutional layers for boring videos are constant in
time, the outputs of pointwise non-linearity layers and average and
max-pooling layers are the same as for the 2D case, and hence the
overall network response respects the boring-video fixed point. \cite{MansimovSS15} studies other bootstrapping strategies.

\vspace{3mm}
\noindent \textbf{Pacing receptive field growth in space, time and network depth.} The boring video fixed-point leaves ample freedom on how to inflate pooling operators along the time dimension and on how to set convolutional/pooling temporal stride -- these are the primary factors that shape the size of feature receptive fields. Virtually all image models treat the two spatial dimensions (horizontal and vertical) equally -- pooling kernels and strides are the same. This is quite natural and means that features deeper in the networks are equally affected by image locations increasingly far away in both dimensions. A symmetric receptive field is however not necessarily optimal when also considering time -- this should depend on frame rate and image dimensions. If it grows too quickly in time relative to space, it may conflate edges from different objects breaking early feature detection, while if it grows too slowly, it may not capture scene dynamics well.

\begin{figure*}
\centering
    \includegraphics[width=1.0\textwidth]{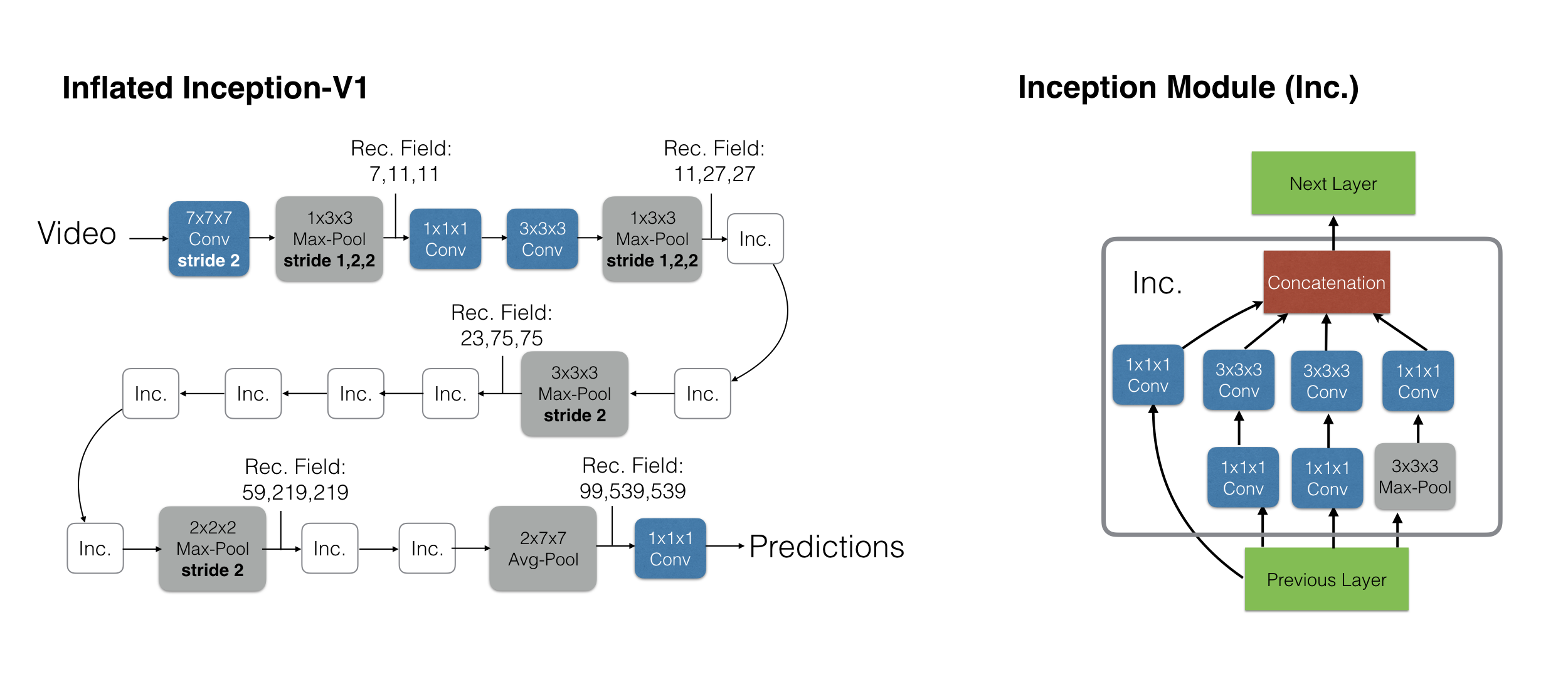}\hspace{3mm}
    \caption{The Inflated Inception-V1 architecture (left) and its detailed inception submodule (right). The strides of convolution and pooling operators are $1$ where not specified, and batch normalization layers, ReLu's and the softmax at the end are not shown.  The theoretical sizes of receptive field sizes for a few layers in the network are provided in the format ``time,x,y" -- the units are frames and pixels. The predictions are obtained convolutionally in time and averaged.}  \label{fig:inception}
\end{figure*}

In Inception-v1, the first convolutional layer has stride $2$, then
there are four max-pooling layers with stride $2$ and a $7 \times 7$
average-pooling layer preceding the last linear classification layer,
besides the max-pooling layers in parallel Inception branches. In our
experiments the input videos were processed at $25$ frames per second;
we found it helpful to not perform temporal pooling in the first two
max-pooling layers (by using $1 \times 3 \times 3$ kernels and stride $1$ in
time), while having symmetric kernels and strides in all other
max-pooling layers. The final average pooling layer uses a $2 \times 7
\times 7$ kernel. The overall architecture is shown in fig.~\ref{fig:inception}. We train the model using $64$-frame snippets and
test using the whole videos, averaging predictions temporally.

\vspace{3mm}
\noindent \textbf{Two 3D Streams.}  While a 3D ConvNet should be able to learn motion features from RGB inputs directly, it still performs pure feedforward computation, whereas optical flow algorithms are in some sense recurrent (e.g. they perform iterative optimization for the flow fields). Perhaps because of this lack of recurrence, experimentally we still found it valuable to have a two-stream configuration -- shown in fig.~\ref{fig:architectures}, e) -- with one I3D network trained on RGB inputs, and another on flow inputs which carry optimized, smooth flow information. We trained the two networks separately and averaged their predictions at test time.

\begin{table*}[ht]
\centering
\begin{tabular}{| c | c | c | c | c | c| }
  \hline
  
  \multirow{2}{*}{\parbox{1.5cm}{Method}} & \multirow{2}{*}{\parbox{1.5cm}{\#Params}} &
  \multicolumn{2}{|c|}{Training} &
  \multicolumn{2}{|c|}{Testing} \\ \cline{3-6}

   & & \# Input Frames & Temporal Footprint & \# Input Frames & Temporal Footprint  \\ \hline \hline
   
ConvNet+LSTM & ~9M & 25 rgb  &  5s &  50 rgb  & 10s \\ \hline
3D-ConvNet & 79M & 16 rgb & 0.64s & 240 rgb & 9.6s \\ \hline
Two-Stream  & 12M & 1 rgb, 10 flow  & 0.4s  & 25 rgb, 250 flow & 10s \\ \hline
3D-Fused & 39M & 5 rgb, 50 flow   & 2s & 25 rgb, 250 flow & 10s \\ \hline
Two-Stream I3D & 25M & 64 rgb, 64 flow & 2.56s & 250 rgb, 250 flow & 10s \\ \hline
\end{tabular} 
\vspace{5pt}
\caption{Number of parameters and temporal input sizes of the models.}
\label{tab:temporal_interfaces}
\end{table*}

\subsection{Implementation Details}

All models but the C3D-like 3D ConvNet use ImageNet-pretrained Inception-V1~\cite{ioffe2015batch} as base network. For all architectures we follow each convolutional layer by a batch
normalization~\cite{ioffe2015batch} layer and a ReLU activation
function, except for the last convolutional layers which produce the
class scores for each network.

Training on videos
used standard SGD with momentum set to 0.9 in all cases, with synchronous
parallelization across 32 GPUs for all models except the 3D ConvNets
which receive a large number of input frames and hence require more
GPUs to form large batches -- we used 64 GPUs for these. We trained
models on on Kinetics for 110k steps, with a 10x reduction of
learning rate when validation loss saturated. We tuned the learning rate hyperparameter on the validation set of
Kinetics. Models were trained for up to 5k steps on UCF-101 and HMDB-51 using a similar learning rate adaptation procedure as for Kinetics but using just 16 GPUs. All the models were implemented in 
TensorFlow~\cite{abadi2016tensorflow}.

Data augmentation is known to be of crucial importance for the
performance of deep architectures. During training we used random
cropping both spatially -- resizing the smaller video side to 256
pixels, then randomly cropping a $224 \times 224$ patch -- and
temporally, when picking the starting frame among those early enough
to guarantee a desired number of frames. For shorter videos, we looped the video as many times as necessary to satisfy each model's input interface. We also applied random left-right flipping consistently for each video during
training. During test time the models are applied convolutionally over the whole video taking $224 \times 224$ center crops, and the predictions are averaged. We briefly tried spatially-convolutional testing on the $256 \times 256$ videos, but did not observe improvement. Better performance could be obtained by also considering left-right flipped videos at test time and by adding additional augmentation, such as photometric, during training. We leave this to future work.

We computed optical flow with a TV-L1 algorithm \cite{zach2007duality}.
 
\section{The Kinetics Human Action Video Dataset\label{dataset}}

The Kinetics dataset is focused on human actions (rather than activities or
events). The list of action classes covers: {\em Person Actions
(singular)}, e.g.\ drawing, drinking, laughing, punching; {\em
Person-Person Actions}, e.g.\ hugging, kissing, shaking hands; and,
{\em Person-Object Actions}, e.g.\ opening presents, mowing lawn, washing dishes.  Some actions are fine grained and require temporal reasoning
to distinguish, for example different types of swimming. Other actions
require more emphasis on the object to distinguish, for example
playing different types of wind instruments.  

The dataset has 400 human action classes, with 400 or more clips for
each class, each from a unique video, for a total of 240k training videos. The clips last around 10s, and
there are no untrimmed videos.  The test set consists of 100 clips for
each class.  A full description of the dataset and how it was built is
given in~\cite{kinetics_arxiv}.

\section{Experimental Comparison of Architectures \label{comp_arch}}

\begin{table*}[h]
\begin{center}
    \begin{tabular}{| l | c | c | c || c | c | c || c | c | c |}
    \hline
 &  \multicolumn{3}{|c||}{UCF-101} &  \multicolumn{3}{c||}{HMDB-51}  &  \multicolumn{3}{|c|}{Kinetics} \\ \cline{2-4} \cline{5-7} \cline{8-10}
Architecture            & RGB   & Flow  & RGB + Flow &  RGB   & Flow  & RGB + Flow & RGB   & Flow  & RGB + Flow    \\ \hline \hline
    (a) LSTM           & 81.0 & -- & -- & 36.0 & --  & -- & 63.3  & -- & --      \\ \hline
    (b) 3D-ConvNet           & 51.6 & -- & -- & 24.3 & --  & -- & 56.1 & -- & --         \\ \hline
    (c) Two-Stream          & 83.6  & 85.6  & 91.2  & 43.2  & 56.3  & 58.3 & 62.2 & 52.4 &   65.6 \\ \hline
    (d) 3D-Fused            &  83.2 & 85.8 & 89.3 & 49.2 & 55.5 & 56.8 & -- & -- & 67.2 \\ \hline \hline
    (e) Two-Stream I3D     & \textbf{84.5} & \textbf{90.6}  & \textbf{93.4} & \textbf{49.8} & \textbf{61.9} & \textbf{66.4} & \textbf{71.1} & \textbf{63.4} & \textbf{74.2}        \\ \hline
    \hline
    \end{tabular}
\end{center}
\caption{Architecture comparison: (left) training and testing on split 1 of UCF-101; (middle) training and testing on split 1 of HMDB-51; (right) training and testing on Kinetics. All models are based on ImageNet pre-trained Inception-v1, except 3D-ConvNet, a C3D-like \cite{tran2015learning} model which has a custom architecture and was  trained here from scratch. Note that
the Two-Stream architecture numbers on individual RGB and Flow streams
can be interpreted as a simple baseline which applies a ConvNet
independently on 25 uniformly sampled frames then averages the
predictions.}
\label{fig:archCompPerf}
\end{table*}

\begin{table*}[h]
\begin{center}
    \begin{tabular}{| l | c | c | c || c | c | c  |}
    \hline
 &  \multicolumn{3}{|c||}{ Kinetics} & \multicolumn{3}{c|}{ImageNet then Kinetics}   \\ \cline{2-4} \cline{5-7} 
Architecture         &  RGB   & Flow  & RGB + Flow & RGB   & Flow  & RGB + Flow    \\ \hline \hline
    (a) LSTM           & 53.9 & --  & -- & 63.3  & -- & --      \\ \hline
    (b) 3D-ConvNet           & 56.1 & --  & -- & -- & -- & --         \\ \hline
    (c) Two-Stream       & 57.9  & 49.6  & 62.8 & 62.2 & 52.4 &   65.6 \\ \hline
    (d) 3D-Fused         & -- & -- & 62.7 & -- & -- & 67.2 \\ \hline \hline
    (e) Two-Stream I3D     & \textbf{68.4} (88.0) & \textbf{61.5} (83.4) & \textbf{71.6} (90.0)
                           & \textbf{71.1} (89.3) & \textbf{63.4} (84.9) & \textbf{74.2} (91.3)      \\ \hline
    \hline
    \end{tabular}
\end{center}
\caption{Performance training and testing on Kinetics with and without ImageNet pretraining. Numbers in
brackets () are the Top-5 accuracy, all others are Top-1.}
\label{fig:archCompPerfKinetics}
\end{table*}

In this section we compare the performance of the five architectures described in  section~\ref{sec:architectures} whilst varying the dataset used for training and testing.

Table~\ref{fig:archCompPerf} shows the 
classification accuracy when training and testing on either UCF-101, HMDB-51 or Kinetics. We test on the split~1 test sets of UCF-101 and HMDB-51 and on the held-out test set of Kinetics. There are several noteworthy observations. First, our new I3D models do best in all datasets, with either RGB, flow, or RGB+flow modalities. This is interesting, given its very large number of parameters and that UCF-101 and HMDB-51 are so small, and shows that the benefits of ImageNet pre-training can extend to 3D ConvNets.

Second, the performance of all models is far lower on Kinetics than on UCF-101, an indication of the different levels of difficulty of the two datasets. It is however higher than on HMDB-51; this may be in part due to lack of training data in HMDB-51 but also because this dataset was purposefully built to be hard: many clips have different actions in the exact same scene (e.g. ``drawing sword" examples are taken from same videos as ``sword" and ``sword exercise").  Third, the ranking of the different architectures is mostly consistent.

Additionally, two-stream architectures exhibit superior performance on
all datasets, but the relative value of RGB and flow differs
significantly between Kinetics and the other datasets. The
contribution from flow alone, is slightly higher than that of RGB on
UCF-101, much higher on HMDB-51, and substantially lower on Kinetics.
Visual inspection of the datasets suggests that Kinetics has much
more camera motion which may make the job of the motion stream
harder. The I3D model seems able to get more out of the flow stream
than the other models, however, which can probably be explained by its
much longer temporal receptive field (64 frames vs 10 during training)
and more integrated temporal feature extraction machinery. While it
seems plausible that the RGB stream has more discriminative
information -- we often struggled with our own eyes to discern actions
from flow alone in Kinetics, and this was rarely the case from RGB
-- there may be opportunities for future research on integrating some
form of motion stabilization into these architectures.


We also evaluated the value of training models in Kinetics starting from ImageNet-pretrained weights versus from scratch -- the results are shown in table~\ref{fig:archCompPerfKinetics}. It can be seen that ImageNet pre-training still helps in all cases and this is slightly more noticeable for the RGB streams, as would be expected.

\begin{figure*}
\centering
    \includegraphics[width=1.0\textwidth]{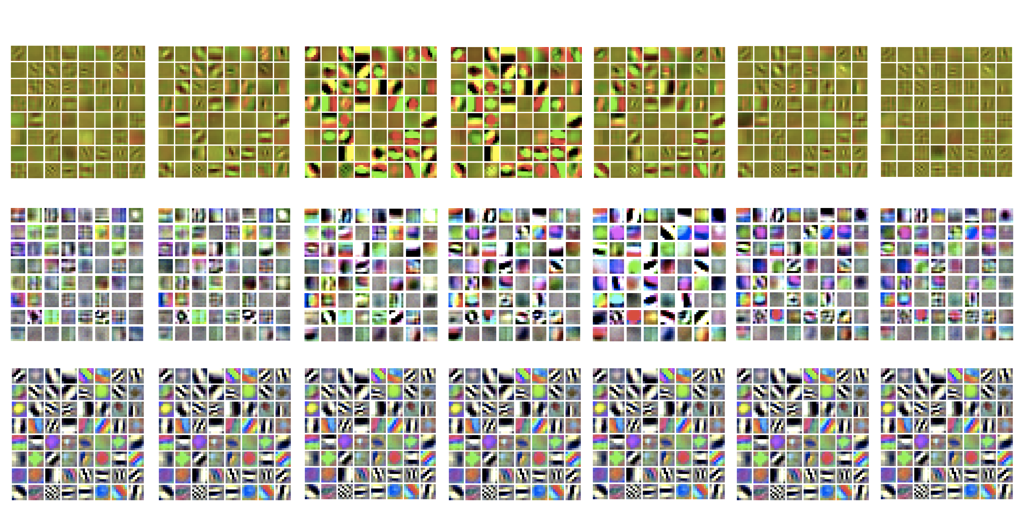}\hspace{3mm}
    \caption{
All 64 conv1 filters of each Inflated 3D ConvNet
after training on Kinetics (the filter dimensions are $7\times7\times7$, and the 7 time dimensions are shown
left-to-right across the figure). The
sequence on top shows the flow network filters, the one in the middle
shows filters from the RGB I3D network, and the bottom row shows the
original Inception-v1 filters. Note that the I3D filters possess rich
temporal structure. Curiously the filters of the flow network are
closer to the original ImageNet-trained Inception-v1 filters, while the filters in the
RGB I3D network are no longer recognizable. Best seen on the computer,
in colour and zoomed in.}  \label{fig:filters}
\end{figure*}

\section{Experimental Evaluation of Features \label{comp_feat}}

In this section we investigate the generalizability of the networks trained on Kinetics. We consider two measures of this: first, we freeze the network weights and use the network to produce features for the (unseen) videos of the UCF-101/HMDB-51 datasets. We then train multi-way soft-max classifiers for the classes of UCF-101/HMDB-51 (using their training data), and evaluate on their test sets; Second, we fine-tune each network for the UCF-101/HMDB-51
classes (using the UCF-101/HMDB-51 training data), and again evaluate on the UCF-101/HMDB-51 test sets.

We also examine how important it is to pre-train on ImageNet+Kinetics instead of just Kinetics. 

The results are given in table~\ref{fig:transferUCFHDMI}. The clear
outcome is that {\em all} architectures benefit from pre-training on
the additional video data of Kinetics, but some benefit
significantly more than the others -- notably the I3D-ConvNet and
3D-ConvNet (although the latter starting from a much lower base). Training just the last layers of the models after pretraining in Kinetics (\textit{Fixed}) also leads to much better performance than directly training on  UCF-101 and HMDB-51 for I3D models.

One explanation for the significant better transferability of features of I3D models is their high temporal resolution -- they are trained on $64$-frame video snippets at $25$ frames per second and process all video frames at test time, which makes it possible for them to capture fine-grained temporal structure of actions. Stated differently, methods with sparser video inputs  may benefit less from training on this large video dataset because, from their perspective, videos do not differ as much from the images in ImageNet. The difference over the C3D-like model can be explained by our I3D models being much deeper, while having much fewer parameters, by leveraging an ImageNet warm-start, by being trained on $4\times $ longer videos, and by operating on $2\times$ higher spatial resolution videos.

The performance of the two-stream  models is surprisingly good even when trained from scratch  (without ImageNet or Kinetics),  mainly due to the accuracy of the flow stream, which seems much less prone to overfitting (not shown). Kinetics pretraining helps significantly more than ImageNet.

\begin{table*}[h]
\begin{center}
    \begin{tabular}{| l | c | c | c || c | c | c|}
    \hline
 &  \multicolumn{3}{|c||}{UCF-101} &  \multicolumn{3}{c|}{HMDB-51}  \\ \cline{2-4} \cline{5-7} 
Architecture & Original   & Fixed & Full-FT & Original   & Fixed & Full-FT  \\ \hline \hline
    (a) LSTM            & 81.0 / 54.2 & 88.1 / 82.6 & 91.0 / 86.8 & 36.0 / 18.3 & 50.8 / 47.1  & 53.4 / 49.7  \\ \hline
    (b) 3D-ConvNet      & -- / 51.6 & -- / 76.0 & -- / 79.9 & -- / 24.3 & -- / 47.0  & -- / 49.4            \\ \hline
    (c) Two-Stream     & 91.2 / 83.6 & 93.9 / 93.3 & 94.2 / 93.8 & 58.3 /  47.1 & 66.6 / 65.9 &  66.6 / 64.3 \\ \hline
    (d) 3D-Fused       & 89.3 / 69.5 & 94.3 / 89.8 & 94.2 / 91.5 & 56.8 / 37.3 & 69.9 / 64.6 & 71.0 / 66.5           \\ \hline \hline
    (e) Two-Stream I3D & 93.4 / 88.8 & 97.7 / 97.4 & 98.0 / 97.6 & 66.4 / 62.2 & 79.7 / 78.6 &  81.2 / 81.3    \\ \hline
    \hline
    \end{tabular}
\end{center}
\caption{Performance on the UCF-101 and HMDB-51 test sets (split 1 of both) for architectures starting with / without ImageNet pretrained weights. Original: train on UCF-101 or HMDB-51; Fixed: features from Kinetics, with the last layer trained on UCF-101 or HMDB-51; Full-FT: Kinetics pre-training with end-to-end fine-tuning on UCF-101 or HMDB-51.}
\label{fig:transferUCFHDMI}
\end{table*}

\subsection{Comparison with the State-of-the-Art}

We show a comparison of the performance of I3D models and previous
state-of-the-art methods in table \ref{fig:state-of-the-art}, on
UCF-101 and HMDB-51. We include results when pre-training on
the Kinetics dataset (with and without ImageNet pre-training). The conv1 filters of
the trained models are shown in fig.~\ref{fig:filters}.

\begin{table*}[h]
\begin{center}
    \begin{tabular}{| l | c | c |}
    \hline
    Model & UCF-101 & HMDB-51  \\ \hline\hline
    Two-Stream~\cite{simonyan2014two} & 88.0 & 59.4 \\ \hline
    IDT~\cite{wangICCV13} & 86.4 & 61.7 \\ \hline
    Dynamic Image Networks + IDT~\cite{bilen2016dynamic} & 89.1 & 65.2 \\ \hline
    TDD + IDT~\cite{wang2015action} & 91.5 & 65.9 \\ \hline
    Two-Stream Fusion + IDT~\cite{feichtenhofer2016convolutional} & 93.5 & 69.2\\ \hline
    Temporal Segment Networks~\cite{wang2016temporal} & 94.2  &  69.4 \\ \hline
    ST-ResNet + IDT~\cite{feichtenhofer2016spatiotemporal} & 94.6 & 70.3  \\ \hline 
    \hline
    Deep Networks~\cite{karpathy2014large}, Sports 1M pre-training & 65.2 & -  \\ \hline
    C3D one network~\cite{tran2015learning}, Sports 1M pre-training & 82.3 & -  \\ \hline
    C3D ensemble~\cite{tran2015learning}, Sports 1M pre-training  & 85.2 & - \\ \hline
    C3D ensemble + IDT~\cite{tran2015learning}, Sports 1M pre-training  & 90.1 & - \\ \hline
    \hline
    RGB-I3D,     Imagenet+Kinetics pre-training & 95.6 & 74.8   \\ \hline
    Flow-I3D,     Imagenet+Kinetics pre-training & 96.7 & 77.1   \\ \hline
    Two-Stream I3D,     Imagenet+Kinetics pre-training & \textbf{98.0} & 80.7   \\ \hline
    RGB-I3D,     Kinetics pre-training & 95.1 & 74.3   \\ \hline
    Flow-I3D,     Kinetics pre-training & 96.5 & 77.3   \\ \hline
    Two-Stream I3D,     Kinetics pre-training & 97.8 & \textbf{80.9}   \\ \hline

    \hline
    \end{tabular}
\end{center}
\caption{
Comparison with state-of-the-art on the  UCF-101 and HMDB-51 datasets, averaged over three splits. First set of rows
contains results of models trained without labeled external data.}
\label{fig:state-of-the-art}
\end{table*}

Many methods get similar results, but the best performing method on these datasets is
currently the one by Feichtenhofer and colleagues~\cite{feichtenhofer2016spatiotemporal}, which uses ResNet-50 models on RGB and optical flow streams, and  gets $94.6$\% on UCF-101 and $70.3$\% on HMDB-51 when combined with the dense trajectories model \cite{wangICCV13}. We
benchmarked our methods using the mean accuracy over the three standard train/test splits. Either of our RGB-I3D or RGB-Flow models alone, when pre-trained on Kinetics, outperforms all previous published performance by any model or model combinations. Our combined two-stream architecture widens the advantage over previous models considerably, bringing overall performance to $98.0$ on UCF-101 and $80.9$ on HMDB-51, which correspond to $63$\% and $35$\% misclassification reductions, respectively compared to the best previous model~\cite{feichtenhofer2016spatiotemporal}.

The difference between Kinetics pre-trained I3D models and prior 3D ConvNets (C3D) is even larger, although C3D is trained on more videos, 1M examples from Sports-1M plus an internal dataset, and even when ensembled and combined with IDT. This may be explainable by the better quality of Kinetics but also because of I3D simply being a better architecture.

\section{Discussion}

We return to the question posed in the introduction, ``is there a
benefit in transfer learning from videos?''.  It is evident that there is a {\em
considerable} benefit in pre-training on (the large video dataset) Kinetics, just as there has been such benefits in pre-training ConvNets on ImageNet for so many tasks. This demonstrates transfer learning from one dataset (Kinetics) to another dataset (UCF-101/HMDB-51) for a similar task (albeit for different action classes). However, it still remains to be seen if there is a benefit in using Kinetics pre-training for other video tasks such as semantic video segmentation, video object detection, or optical flow computation. We plan to make publicly available I3D models trained on the official Kinetics dataset's release to facilitate research in this area. 

Of course, we did not perform a comprehensive exploration of architectures --
for example we have not employed action
tubes~\cite{gkioxari2015finding,Klaser10} or attention 
mechanisms~\cite{li2016videolstm} to focus in on the human actors. Recent works have proposed imaginative methods for determining the spatial and temporal extent (detection) of actors within the two-stream
architectures, by incorporating linked object detections in
time~\cite{peng2016multi,Saha_2016}.  The relationship between space and time is a mysterious one. Several
very creative papers have recently gone out of the box in attempts to
capture this relationship, for example by learning
frame ranking functions for action classes and using these as a
representation~\cite{fernando2015modeling}, by making analogies
between actions and transformations~\cite{wang2015actions}, or by
creating 2D visual snapshots of frame
sequences~\cite{bilen2016dynamic} -- this idea is related to the
classic motion history work of~\cite{bobick2001recognition}. It would
be of great value to also include these models in our comparison but
we could not, due to lack of time and space.

\subsection*{Acknowledgements:} We would like to thank everyone on the Kinetics project and in particular Brian Zhang and Tim Green for help setting up the data for our experiments, and Karen Simonyan for helpful discussions.

{\small
\bibliographystyle{ieee}
\bibliography{references}
}

\newpage

\end{document}